\documentclass[twocolumn,fleqn,natbib]{svjour2}
\smartqed  % flush right qed marks, e.g. at end of proof
\usepackage{amsmath,amsfonts,amssymb}
\usepackage{graphicx}
\usepackage{setspace}
\usepackage{tocloft}
\usepackage{caption}
\usepackage{subfig}
\usepackage{makecell}
\journalname{Real-Time Image Processing}
\begin{document}

\title{Real-Time Optical flow-based Video Stabilization for Unmanned Aerial Vehicles}
%\subtitle{Do you have a subtitle?\\ If so, write it here}

%\titlerunning{}        % if too long for running head

\author{Anli Lim \and
		Bharath Ramesh \and
        Yue Yang \and
        Cheng Xiang \and
        Zhi Gao \and
        Feng Lin
}

%\authorrunning{Short form of author list} % if too long for running head

\institute{Anli Lim, Bharath Ramesh, Yue Yang and Cheng Xiang \at
           Department of Electrical and Computer Engineering    \\
           National University of Singapore \\
           Singapore 117576 \\
              Tel.: +65 65164258\\
              Fax: +65 67791103\\
              \email{bharath.ramesh03@u.nus.edu}           %  \\
%             \emph{Present address:} of F. Author  %  if needed
           \and
           Zhi Gao and Feng Lin \at
           Temasek Laboratories \\
           National University of Singapore \\
           Singapore 117576 \\
}

\date{Received: date / Revised: date}
% The correct dates will be entered by the editor

\maketitle

\begin{abstract}
This paper describes the development of a novel algorithm to tackle the problem of real-time video stabilization for unmanned aerial vehicles (UAVs). There are two main components in the algorithm: (1) By designing a suitable model for the global motion of UAV, the proposed algorithm avoids the necessity of estimating the most general motion model, projective transformation, and considers simpler motion models, such as rigid transformation and similarity transformation. (2) To achieve a high processing speed, optical-flow based tracking  is employed in lieu of conventional tracking and matching methods used by state-of-the-art algorithms. These two new ideas resulted in a real-time stabilization algorithm, developed over two phases. Stage I considers processing the whole sequence of frames in the video while achieving an average processing speed of 50fps on several publicly available benchmark videos. Next, Stage II undertakes the task of real-time video stabilization using a multi-threading implementation of the algorithm designed in Stage I.
\end{abstract}

\section{Introduction}
\label{intro}
In recent times, unmanned aerial vehicles (UAVs) are often equipped with streaming video cameras that can be employed for immediate observation. They are popular for several applications such as rescue, surveillance, mapping, etc. The main drawbacks of videos taken by UAV are the undesired shaking motion caused by atmospheric turbulence and jittery flight control of the platform. The shaky motion in the video proves to mitigate the fundamental aim of using UAV for vision-based tasks. In particular, the unstable motion also inhibits higher level vision tasks, such as detecting and tracking of objects in the video. While significant work has been done for handheld cameras and ground vehicles, there are limited works for stabilization of videos taken from UAV in the literature. Moreover, the existing UAV video stabilization algorithms, cannot be applied on-board UAV systems in real-time.

\begin{figure}
\centering
\includegraphics[width=3.5in]{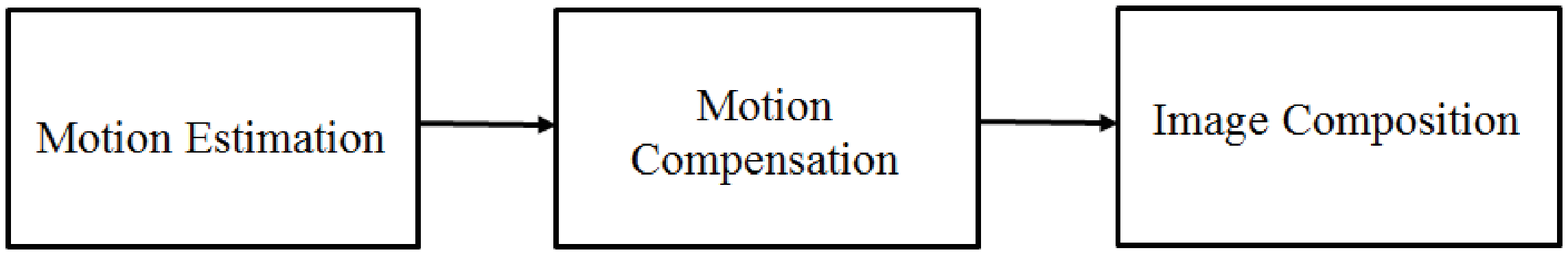}
\caption{Video Stabilization Framework}
\label{figure:VSWorkFlow}
\end{figure}

In general, UAV video stabilization algorithms (\citet{shen2009fast,vazquez2009real,wang2011real,dong2014instantaneous,lucas1981iterative}) follow three main steps: (1) motion estimation, (2) motion compensation and (3) image composition as indicated in Figure \ref{figure:VSWorkFlow}. Most methods revolve around finding the 2D motion model (such as homography) to estimate the global motion trajectory. Then a low-pass filter is applied to the trajectory to sieve out the high frequency jitters. Then the low frequency parameters are then applied onto frames via warping. This framework is really effective for scenes with very little dynamic movement which is applicable to aerial videos taken by UAVs. However, estimating the global motion trajectory using a window of frames cannot achieve instantaneous stabilization, as required for real-time processing.

In \cite{shen2009fast}, a video stabilization algorithm for UAV was proposed using a circular block to search and match key places. The estimated affine transform was then smoothed by the polynomial fitting and prediction method (PFPM). However, this approach only achieved a speed of less than 10fps on a desktop with 3.0GHz processor and 1GB RAM for images with resolution of 216x300p. 

In \cite{vazquez2009real}, a smoothing method utilizes Lucas-Kanade tracker(\cite{lucas1981iterative}) to detect interest points. The unintended motion compensation was accomplished by adjusting for extra rotation and displacements that generate vibrations. This approach is able to achieve a stabilizing speed between 20fps and 28fps for images with resolution of 320x240 pixels on a MAC laptop with 2.16GHz Intel Core 2 Duo Processor with a three frame delay. 

\cite{wang2011real}, later proposed a three-step video stabilization method for UAVs. Firstly, a FAST corner detector is employed to locate the feature points in the frames. Secondly, the matched key-points are used for estimation of affine transform to reduce false matches. Finally, motion estimation is performed based on the affine model and the compensation for vibration is conducted based on spline smoothing. It was reported that this algorithm can process up to 30fps on a Workstation with an Intel Xeon 2.26 GHz processor and 6 GB RAM for images with resolution of 320×240 pixels.

A very recent work by \cite{dong2014instantaneous} is able to process a 640x480 pixels video at a speed of 40fps on a notebook computer with a 2.5GHz Intel Duo Core CPU. While it is able to achieve a high processing speed, the improvement comes at the price of unreliable motion estimation, which leads to failure in stabilizing the video ultimately. Moreover, the algorithm proposed does not furnish evidence for real-time processing on real-time on-board systems. 

It is noted that these previous methods work offline or in a post-process way. For some UAV vision tasks, frames are required to be stabilized immediately and be presented to other tasks, such as object tracking and detection.

The proposed method in this paper is closely related to the method proposed by Ho (\cite{nghiaho2014simple}). The algorithm employs finding corners in frames, followed by estimating a 2D motion model between consecutive frames. Then the parameters in motion model are treated as trajectory to be smoothed using an averaging window. This smoothing of the motion model parameters is different from the motion trajectory smoothing employed by previous works (\cite{shen2009fast,vazquez2009real,wang2011real,dong2014instantaneous,lucas1981iterative}). Our framework is similar to Nghia Ho's but has a significant difference in terms of implementation and performance. Most notably, a novel hybrid mechanism for motion estimation and an optical flow-based corner tracker has been proposed to overcome the challenges encountered by previous algorithms. In addition, the proposed stabilization algorithm performs in real-time, whereas Nghia Ho's algorithm processes the whole video sequence before achieving stabilization.
 
The rest of this paper is structured as follows. In section 2, we introduce the proposed framework for video stabilization. Then in section 3, we discuss how the processing speed of motion estimation is expedited by using an optical flow-based corner tracker and a reduced region of interest. Then in section 4, we explain the improvisation of the motion estimation given by optical flow-based tracker with hybrid motion estimation mechanism. Following which, we verify the speed of the algorithm on publicly available UAV videos and achieve a speed faster than the current state-of-the-art algorithm. In section 5, we discuss the implementation of the real-time component of the algorithm using multi-thread processing. Finally, we conclude the paper in section 6.

\section{Proposed Framework}
\begin{figure*}
\centering
\includegraphics[width=6in]{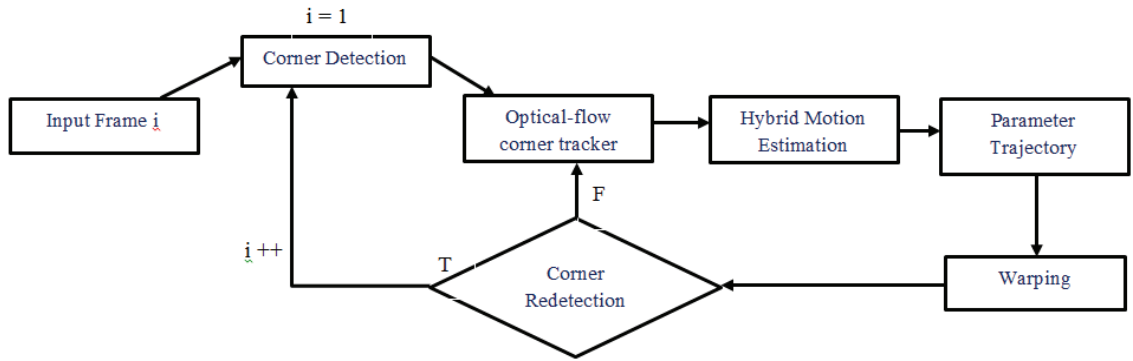}
\caption{Overview of the proposed video stabilization framework}
\label{figure:NovelWorkFlow}
\end{figure*}

Spurred by the challenge to meet the real-time requirements, we propose a novel stabilization framework as shown in Figure \ref{figure:NovelWorkFlow}. The idea is to build an appropriate model for the global motion trajectory of the UAV camera, estimate the motion parameters between two successive frames using very efficient key point detector and compensate the motion via an effective optical flow based tracking of the key points.

Firstly, the key novelty of the proposed algorithm lies in the high speed estimation of the motion parameter trajectory between two consecutive frames, which other methods are less efficient in accomplishing. This is because the motion estimation phase chooses an appropriate lower order homography, either rigid or similarity transformation between frames. This is so that the algorithm is able to frequently use lower order of homography to estimate the motion parameters unless there is a need to use higher order. As a result, achieving a high processing speed. Secondly, our motion compensation phase smooths the parameters via an averaging window. Lastly, we warp the frame according to the smoothed parameters. The program is able to process up to 100 frames per second for all 3 stages of the algorithm. This reassures us that the algorithm is suited for real-time video stabilization.

After achieving such a high speed in stabilizing the frames, the real-time version of the algorithm can be implemented. The three stages of video stabilization were separated into three different threads. Threads are basically independent processes that can run concurrently and share the same resources inside a program. Our algorithm is implemented such that motion estimation thread, motion compensation thread and image composition thread are synchronized in such a way that they are able to stabilize a UAV video in real time. In other words, there is no need to wait for the processing for the whole video to be done.

\section{Optical-Flow-Based Tracking}

Motion estimation is the very first step in video stabilization and it is also the most time consuming step. Consequently, our primary goal is to cut down the time required to calculate the motion parameters trajectory between consecutive frames in the video. The motion trajectory is estimated from how the feature points (\cite{harris1988combined}) move between consecutive frames. Therefore, to initialize motion estimation, we first efficiently detect feature points.

\subsection{Feature Point Detection}

Feature points are locations in the image with large variations in intensity in all directions. They are very important in motion estimation step as they determine the quality of the motion estimation. One early attempt to find these corners was done by Chris Harris and Mike Stephens in their paper (\cite{harris1988combined}), which now is called Harris Corner Detector. It basically finds the difference in intensity for a displacement of (u, v) in all directions. This can be easily expressed as below.

\begin{equation}
E(u,v) = \sum\limits_{x,y} {w(x,y){{[I(x + u,y + v) - I(x,y)]}^2}} 
\label{eq:HarrisCornerDetector}
\end{equation}

where w (x, y) is either a rectangular window or Gaussian window function which gives weight to the surrounding pixels. In 1994, \cite{shi1994good} made a small modification to it in their paper which shows better results compared to Harris Corner Detector. The scoring function in Shi-Tomasi Corner Detector is given by:

\begin{equation}
	R = \min ({\lambda _1},{\lambda _2})
	\label{eq:HarrisScore}
\end{equation}

\begin{itemize}
\item •	When $|R|$  is small, which happens when ${\lambda _1}$ and ${\lambda _2}$ are small, the region is flat.
\item •	When $R$ , which happens when ${\lambda _1}$ $\gg$ ${\lambda _2}$ vice versa, the region is edge.
\item •	When $R$  is large, which happens when ${\lambda _1}$  and ${\lambda _2}$ are large and ${\lambda _1}$ ~ ${\lambda _2}$ , the region is a corner.
\end{itemize}

Since the proposed idea is to ensure that the algorithm is able to run in real-time, a simple experiment of mosaic was conducted. Its purpose is to find out if the number of key points detected has a huge effect on the estimated transformation parameters and also to find out how the time is affected when the number of key points returned is scaled up. 

\begin{figure*}
  \centering
  \begin{tabular}[b]{c}
    \includegraphics[width=.23\linewidth]{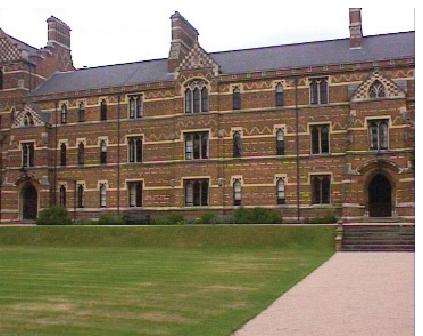} \\
    \small (a)
  \end{tabular}
  \begin{tabular}[b]{c}
    \includegraphics[width=.23\linewidth]{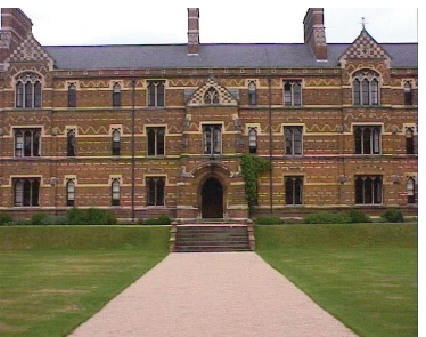} \\
    \small (b)
  \end{tabular}
  \begin{tabular}[b]{c} 
    \includegraphics[width=0.23\linewidth]{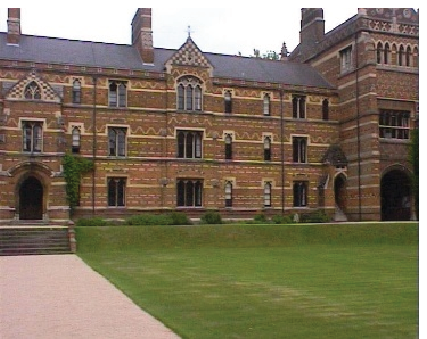} \\
    \small (c)
  \end{tabular}
  \caption{Mosaic Images}
  \label{fig:Mosaic}
\end{figure*}

In this experiment, two images of the same scene will be stitched together with a reference image as shown in Figure \ref{fig:Mosaic}.

Firstly, key points will be detected in the left image and the reference image. Next, key point matching will be performed between the left image and the reference image. Finally, a homography will be generated to transform image 1 to the perspective of the reference image. The cycle repeats itself for the right image.

\begin{figure*}
  \centering
  \begin{tabular}[b]{c}
    \includegraphics[width=.35\linewidth]{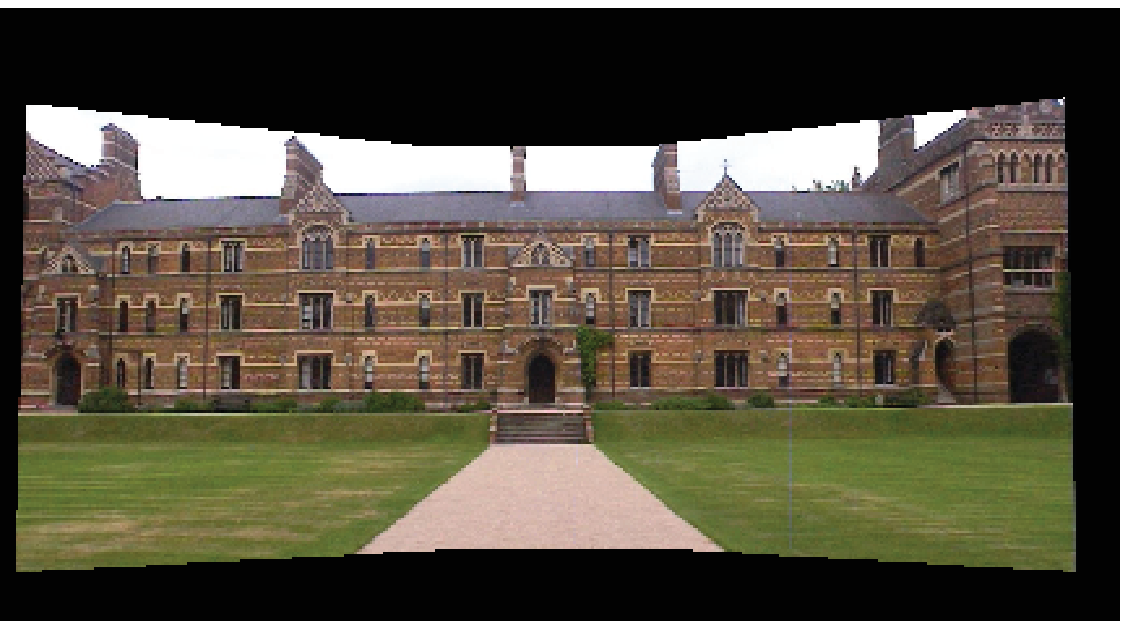} \label{fig:200KPMosaic}\\
    \small (a) 200 Key Points
  \end{tabular}
  \begin{tabular}[b]{c}
    \includegraphics[width=.35\linewidth]{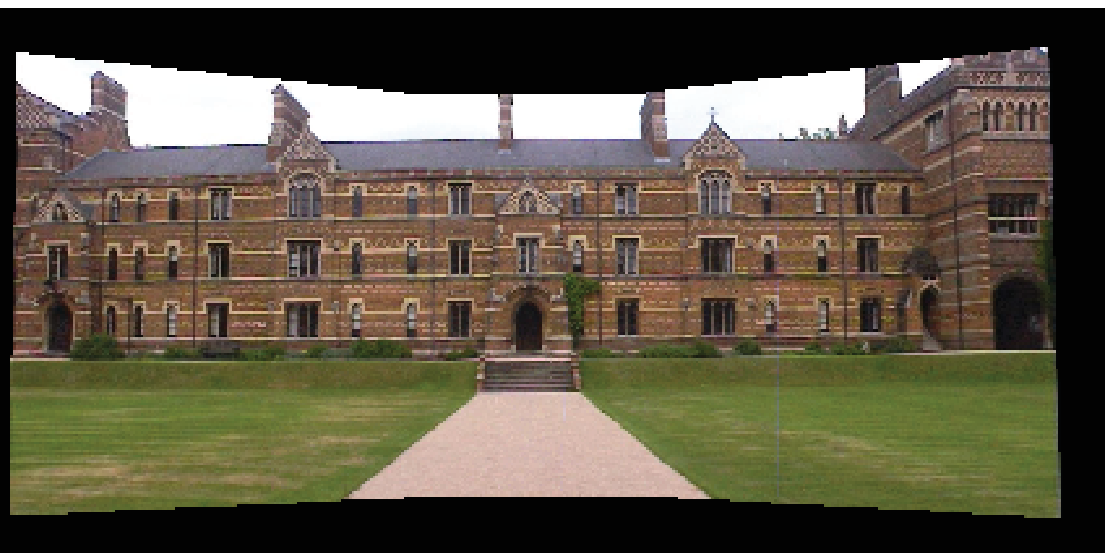} \label{fig:1000KPMosaic} \\
    \small (b) 1000 Key Points
  \end{tabular}
  \caption{Mosaic Results}
  \label{fig:Mosaic Results}
\end{figure*}

If observed carefully, there is not a substantial qualitative difference between a 200 Key Points Mosaic image and a 1000 Key Points Mosaic image as illustrated in Figure \ref{fig:Mosaic Results}. The quality of both mosaics are similar. This demonstrates that large number of Harris corners is unnecessary. Therefore, keeping in mind that our proposed idea is to run the algorithm in real-time, the number of key points needed for a good estimate of the motion model is set to be less than 200. From experiments, we set a value of 50 which gave us visually pleasing results without compromising on stabilization quality.

\begin{figure}
\centering
\includegraphics[width=2in]{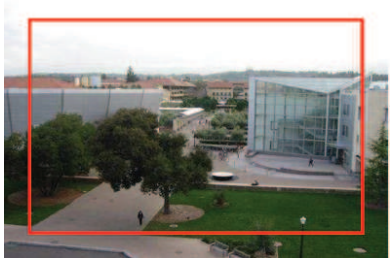}
\caption{Reduced Region of Interest}
\label{figure:ROI}
\end{figure}

Next, the area for detection of points is implemented such that the key point detection will be from an area smaller than that of the frame (see Figure \ref{figure:ROI}). The region of interest will be set such that the length and width of the region are smaller than the size of the frame. The reason why this region is implemented is because the pixels close to the edges of the frame have a high probability of not appearing in the next consecutive frame, especially when the UAV is in motion. In this way, it can prevent the key points at the edges from being detected, saving computational time needed to match those key points which have a high probability of vanishing in the next frame.  The next step is to match the key points.

\subsection{Feature Point Matching}
To speed up the matching, a common way is to use Approximate Nearest Neighbors (\cite{andoni2006near}).This method is still slow for our real-time application. Furthermore, these techniques are only useful when the objective is just to characterize the neighbourhood rather than the exact locations. In image stabilization, locations are important so that the homography generated is accurate. Therefore, we propose an optical flow based matching of corners detected in one frame to the next frame.

Optical flow (\cite{fleet2006optical}) is the pattern of apparent motion of image objects between two consecutive frames caused by the movement of an object or camera. It is a 2D vector field where each vector is a displacement vector showing the motion of points from first frame to second. Optical flow assumes brightness constancy, as shown below. 

\begin{equation}
I(x,y,t) = I(x + dx,y + dy,t + dt)
\label{eq:Optical-flow}
\end{equation}

Then taking taylor series approximation of right-hand side, and removing common terms and dividing by $dt$ to get the following equation:

\begin{equation}
f _x u + f _y v + f_t = 0
\label{eq:Optical-flow1}
\end{equation}

where:

\begin{equation}
f _x = \frac{{\partial { f}}}{{\partial {{x}}}};{{f _y}} = \frac{{\partial {\rm{f}}}}{{\partial {{y}}}}
\label{eq:Optical-flow2}
\end{equation}

\begin{equation}
{{u}}  = \frac{{dx}}{{dt}};v = \frac{{dy}}{{dt}}
\label{eq:Optical-flow3}
\end{equation}

The above equations describe the optical flow in terms of the spatial image gradient. In it, we can find $f _x$ and $f _y$, they are image gradients. Similarly, $f _t$ is the gradient along time, but $(u,v)$ is unknown. We cannot solve this one equation with two unknown variables. Various methods have been suggested to resolve this problem and we choose the gold standard, Lucas-Kanade algorithm (\cite{lucas1981iterative}).

\begin{figure}
	\centering
    \includegraphics[width=3.5in]{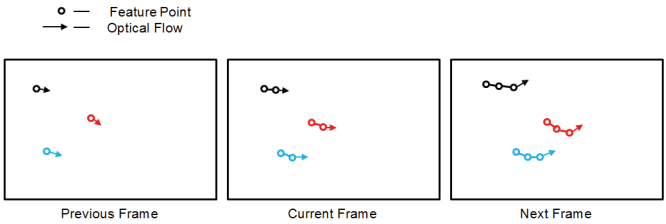}
    \caption{Optical-Flow to estimate motion}
    \label{fig:ME btwn 3 frames}
\end{figure}

By estimating the movement of the corner points, we do not have to detect corners in the next frame. And since we have estimated where the corners are in the next frame, matching is already completed for us. Optical flow matching is repeated for five frames with corner detection only for the first frame, and then the corners are re-detected in the sixth frame. This is done to not violate the two main assumptions of optical flow, which are:

1.	The pixel intensities of an object do not change between consecutive frames.

2.	Neighbouring pixels have similar motion.

Using the Lucas-Kanade method, we obtain the optical flow vectors of the corner points in the first frame and subsequently for the next four frames. But again, there are some problems. Until now, we were dealing with small motions and the assumptions fail when there is large motion. So we use pyramids (\cite{adelson1984pyramid,bouguet2001pyramidal}) to remove small motions and large motions becomes small motions as we go higher in the pyramid. Now applying Lucas-Kanade, we get the optical flow along with the scale. 

We detect the Shi-Tomasi corner points in the first frame. Then, we track those points using Lucas-Kanade optical-flow in iteration. In subsequent frames, the tracker will provide us with the new locations of the corner points. We pass these next points as the previous points in the next step. 

The motion vectors are the coordinates of the key points that existed in the previous frame and have moved to a new location in the current frame. After retrieving the new key point locations, we know that there exists a set of corners that are both in the previous frame and the current frame. This permits us to generate an inter-frame transformation matrix that is used for motion modeling and image composition.

\section{Homography Estimation}
After acquiring the corners that exist in both current and previous frames, we can now estimate the motion model between these frames. As mentioned above, we know the pixels in the previous frame have moved in the current frame. Using this information, the motion estimator takes in the coordinates of the feature points in both frames to generate a homography. This function allows the homography returned to be of similarity (also known as partial affine) or rigid (also known as Euclidean) transformation and it is implemented such that the homography returned depends on the hybrid motion estimation used. The deciding factor of the mechanism will be discussed in the next following subsection. If the homography returned is a rigid transformation, the transform parameters will store only the tx, ty, angle of the transformation and a scale factor of 1. If the homography matrix returned is of similarity transformation, the transform parameters will store tx, ty, angle and an arbitrary scale factor of the transformation. After we extract these transform values, we treat them as motion parameters.

\subsection{Hybrid mechanism for motion estimation}

In previous works, motion estimation is carried out either using particle filters (\cite{yang2006online}) , or variational methods (\cite{pilu2004video}) , and mostly using SIFT feature based matching techniques(\cite{battiato2007sift}) , all of which are not suitable for the real-time implementation considered in this work. An extensive discussion of this point can be found in (\cite{dong2014instantaneous}). In fact, Dong, pointed out that the requirement of the long-range feature tracking makes many existing methods incompetent for challenging cases, since long feature trajectories are difficult to obtain in sequences with rapid scene changes, texture less objects, severe occlusions, or excessive motion blur. All these scenarios are common for UAV videos and they are tackled in this work. Citing the above reasons, Dong made use of a fast KLT tracker and achieved an average speed of 50 fps on publicly available UAV videos. 

In general, the motion estimation finds an affine transform [A|t] (a 2 x 3 floating-point matrix) that approximates best the affine transformation between two sets of points. In case of feature point sets, the problem is formulated as follows: you need to find a 2x2 matrix A and 2x1 vector t for the source points (src) and the destination points (dst).

\begin{equation}
[{A^*}|{t^*}] = \arg \min \sum\limits_i {||dst[i] - Asrc{{[i]}^T} - t|{|^2}}
\label{eq:Homography}
\end{equation}

Solving for [A|t]  requires a minimum of 3 pairing points that are not degenerate. This is straight forward to do. Let’s denote the src point to be X= [x y 1] and the dst to be Y = [x’ y’ 1], giving:

\[TX = Y\]
\begin{equation}
\left[ \begin{array}{l}
\begin{array}{*{20}{c}}
{{a_{11}}}&{{a_{12}}}&{{t_x}}
\end{array}\\
\begin{array}{*{20}{c}}
{{a_{21}}}&{{a_{22}}}&{{t_y}}
\end{array}
\end{array} \right]\left[ {\begin{array}{*{20}{c}}
x\\
y\\
1
\end{array}} \right] = \left[ {\begin{array}{*{20}{c}}
{x'}\\
{y'}\\
1
\end{array}} \right]
\label{eq:Homography1}
\end{equation}

We can re-write this as a typical Az = b matrix and solve for z. We’ll also need to introduce 2 extra pair of points to be able to solve for the motion parameters, as follows.

The affine transform mentioned earlier has a degree of freedom of six. There exist two lower degree of freedom transformations, namely similarity and rigid transformation. Those two transformations have four and three degrees of freedom respectively. The four degrees of freedom include rotation angle, scaling, translation in x and translation in y. Rigid transformation assumes no scaling.

\begin{figure*}
\centering
\includegraphics[width=5in]{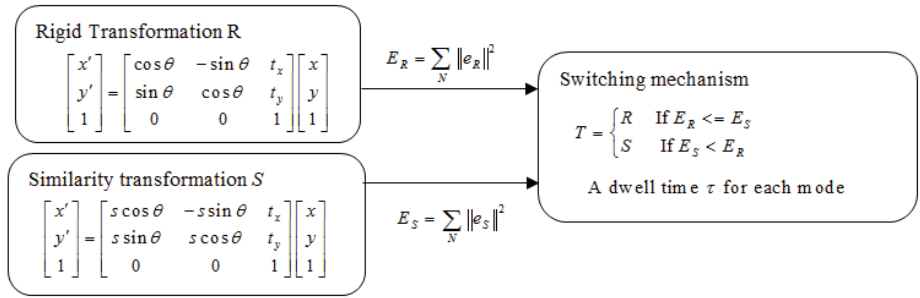}
\caption{Hybrid Mechanism for Motion Estimation}
\label{figure:SwitchingMechanism}
\end{figure*}

Figure \ref{figure:SwitchingMechanism} illustrates the idea of the hybrid mechanism. $E _r$ and $E _s$ represent the total root-mean-square distance, ($e _r$, $e _s$), between the $N$ corners of the previous frame and current frame, after applying the respective transformations. These two values will be computed and compared. If the root-mean-square distance for rigid transformation is lower than similarity transformation, then the mechanism will select rigid transformation computation for a sequence of frames (i.e. dwell time). To achieve a balance between speed and stability, a dwell time of 20 frames is used to switch between rigid (3 parameters) and similarity transformation (4 parameters).

For every 20 frames, when this mechanism is used, key points that are detected will be used together with the two homography matrices, namely rigid transformation matrix and similarity transformation matrix. At the initial frame both rigid and affine matrices are estimated from the set of key points that are detected and matched from optical flow. Then, we estimate two sets of new motion vectors using the two transformation matrices. The next step in the algorithm will then calculate the average distance between the new points and the key point locations for both the transformations. If the average distance between the points generated for rigid transformation is smaller than that of the average distance between the points generated for similarity transformation, then the mechanism will remain to be rigid transformation,  or switch to similarity transformation. Therefore, the algorithm will return the transformation that gives lesser change in distance between the points.

The hybrid mechanism is switching between rigid and partial affine transformation. This means that we are stabilizing shaky motions using only up to 4 degrees of freedom, namely, translation in X direction, translation in Y direction, the scale factor and the rotation factor. 

Based on the video stabilization algorithm described so far, we have accomplished the implementation of the homography hybrid mechanism. The hybrid  mechanism has achieved two main objectives:

\begin{enumerate}
\item •	Reduce movement of the features, thereby reducing jitters between two successive frames. 
\item •	Decrease computational time by choosing the transformation with less parameters whenever possible.
\end{enumerate}

\begin{figure*}
\centering
\includegraphics[width=4.5in]{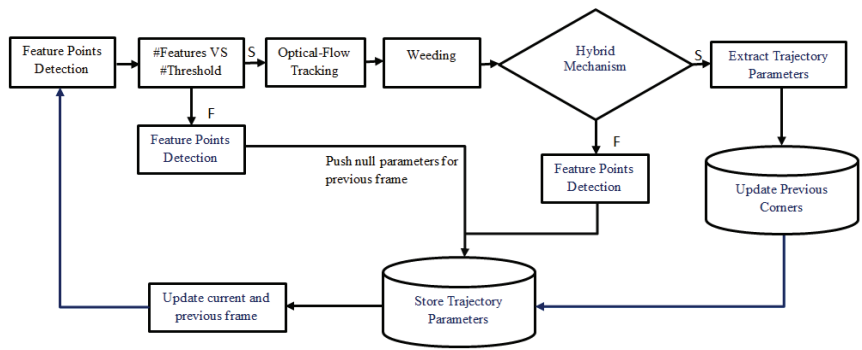}
\caption{Motion Estimation}
\label{figure:motionestimation}
\end{figure*}

Figure \ref{figure:motionestimation} gives an overview of motion estimation work flow in the proposed algorithm. It begins by detecting corners in frames, then utilizes optical-flow to detect and track corners in subsequent frames. On a rare basis, if the number of corners detected is too less, then stabilization is skipped for that frame. The optical flow tracking is followed by weeding step which checks for flow consistency in the backward direction, thereby eliminating bad matches. For every twenty frames, hybrid mechanism is called upon to determine the best transformation to be estimated. Then the parameter trajectories are extracted from the estimation and stored. The process is repeated for the number of frames in the video.

\section{Motion Compensation and Image Composition}
\subsection{Averaging Window Smoothing}

In the motion compensation stage, we employ a basic method of accumulating trajectory parameters and then smoothing them using an averaging window. The trajectory of the parameters is accumulated within a window of frames, which is double the size of smoothing radius, and then the result of the addition of each transformation parameters is then averaged out according to the window size. The result of the averaging of the parameters is then the smoothed trajectory. Next, iterate through each of the original trajectory and adding the difference between the smoothed trajectory and the accumulated global trajectory. 

\begin{figure*}
  \centering
  \begin{tabular}[b]{c}
    \includegraphics[width=.35\linewidth]{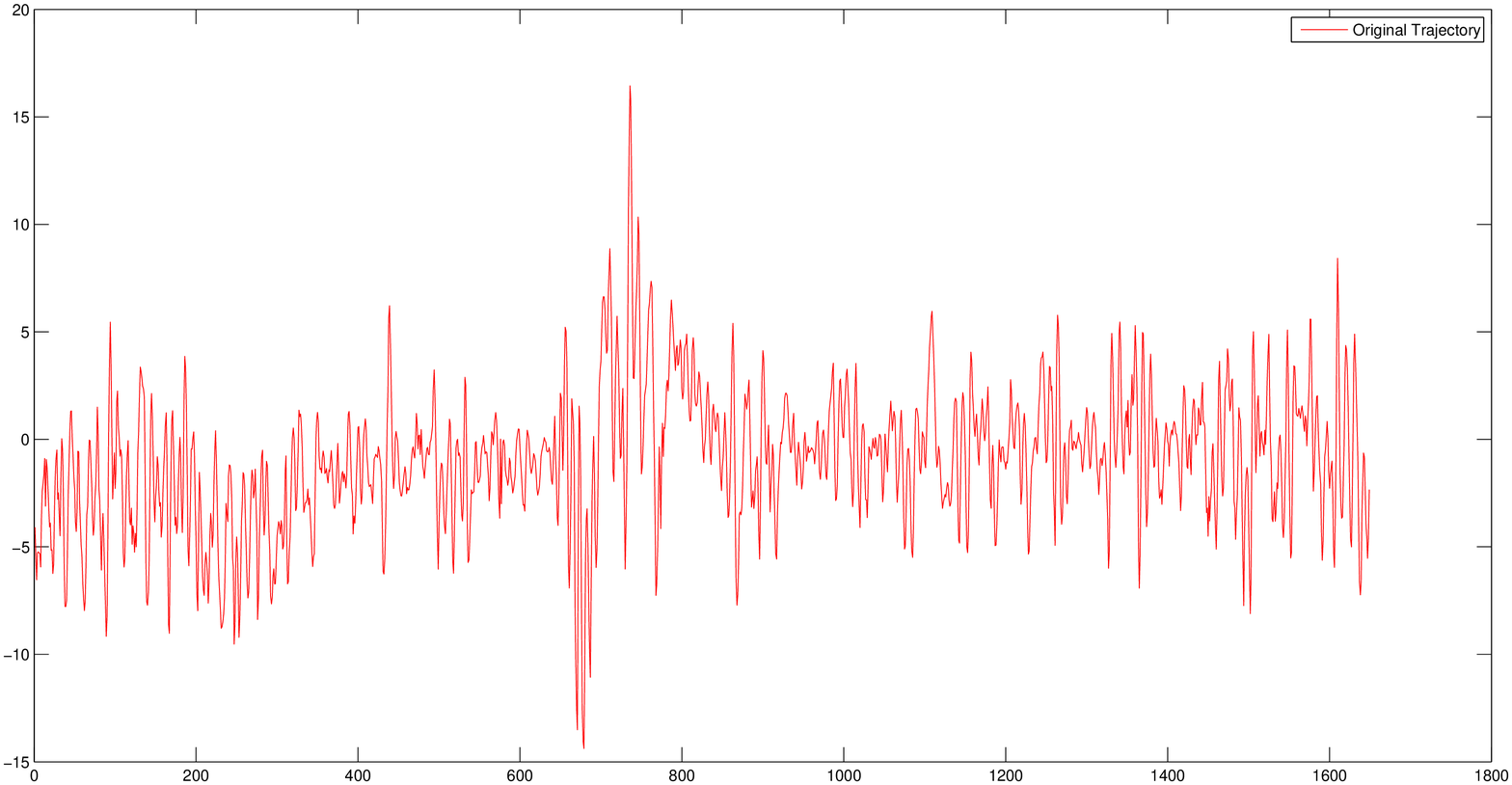} \\
    \small (a) Original Trajectory
  \end{tabular}
  \begin{tabular}[b]{c}
    \includegraphics[width=.35\linewidth]{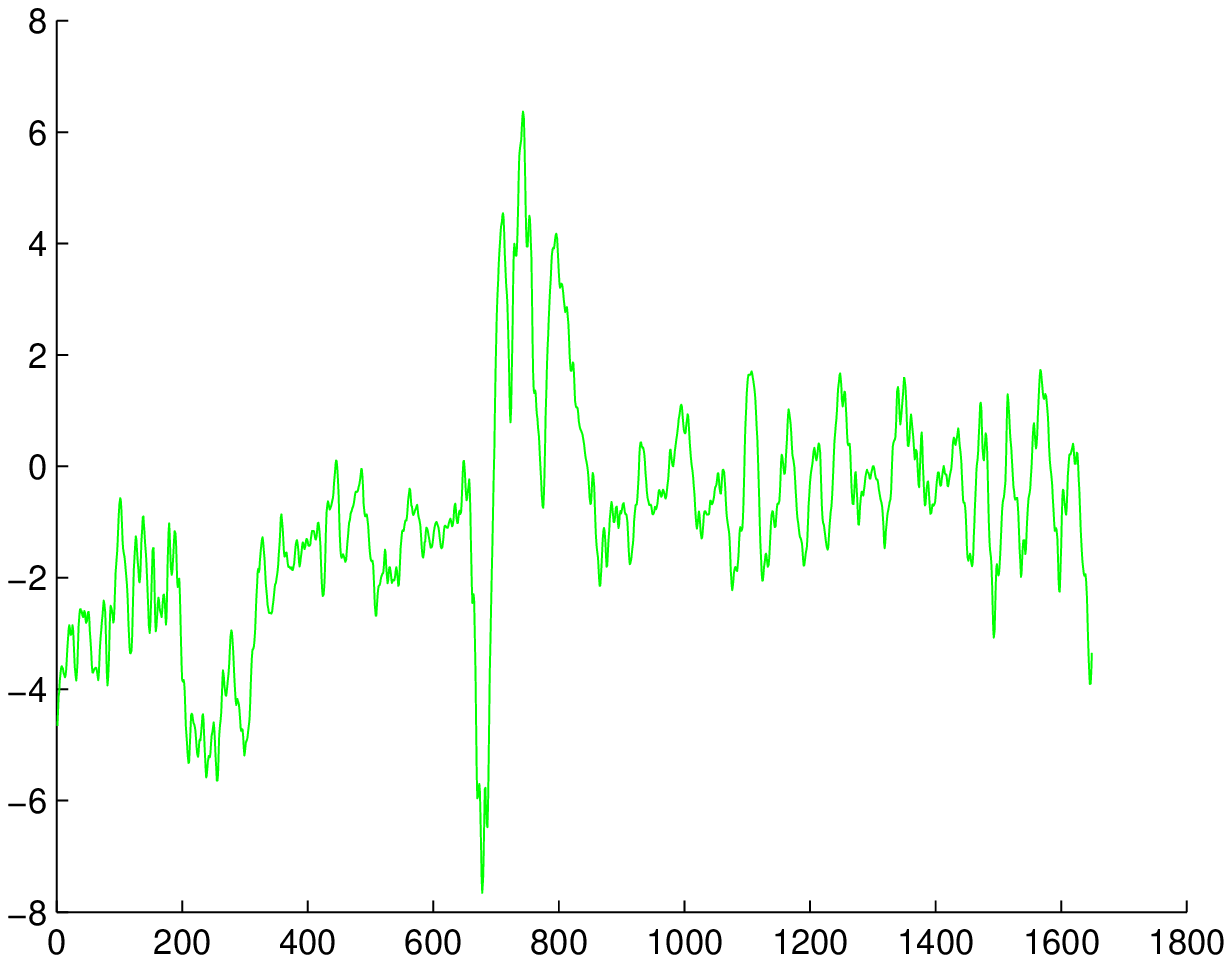} \\
    \small (b) Smoothed Trajectory
  \end{tabular}
  \caption{Compensation Results}
  \label{fig:Smoothed_Trajectory}
\end{figure*}

Figure (a) of \ref{fig:Smoothed_Trajectory} shows the values of $t _x$. It can be viewed in Figure (b) of \ref{fig:Smoothed_Trajectory} that the values of $t _x$ fluctuate a lot, which correlates with our observation of the video being shaky. The next step is to smooth the x values separately after accumulating the frame-to-frame transformation using a sliding average window. It is worth mentioning that the parameter trajectory is a rather abstract quantity that doesn’t necessarily have a direct relationship to the motion induced by the camera. For a simple panning scene with static objects, it has a direct relationship with the absolute position of the image. The important insight is that the trajectory can be smoothed, even if it does not have any physical interpretation. The result of the smoothing process is shown in Figure \ref{fig:Smoothed_Trajectory}.

\begin{figure*}
\centering
\includegraphics[width=6in]{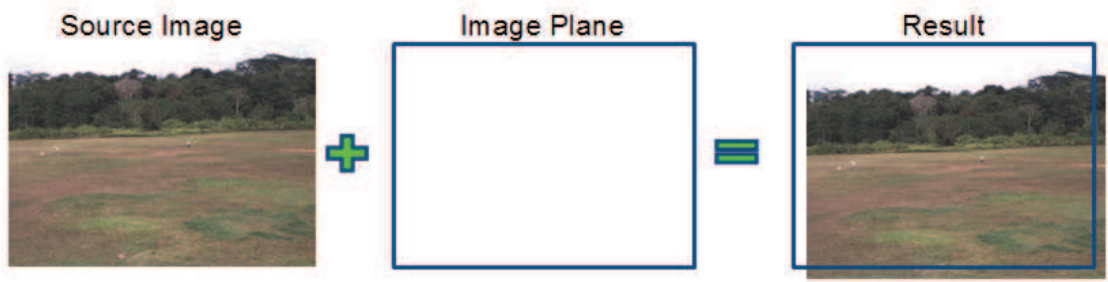}
\caption{Image Warping into Image Plane}
\label{figure:Forward Warping}
\end{figure*}

In the final step of the video stabilization algorithm, a basic method is used to couple warping with cropping. First, warp the frame according to the new set of homography derived from the previous step. Next, the cropping and resizing will reduce unfilled area cause by the smooth motion parameters. 

\section{Real-Time Implementation}
\label{6realtime}
In the previous sections, it is discussed that the algorithm works by these 3 steps, motion estimation, motion compensation and image composition. These 3 steps are usually executed one after another. This means that the algorithm has to have the entire video as the input, then it will process all the frames within the video via those steps mentioned and then display the output, which is not real-time at all.

\begin{figure*}
\centering
\includegraphics[width=5in]{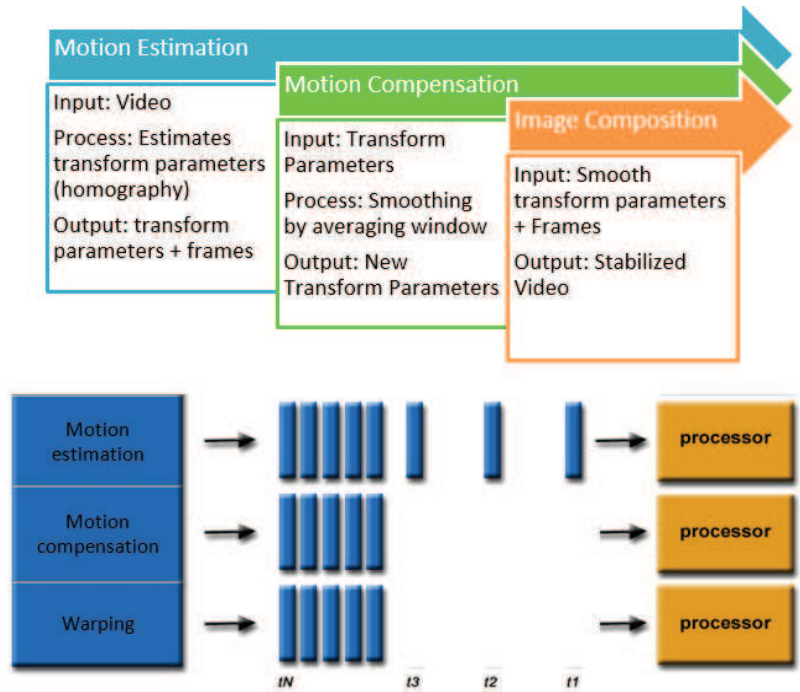}
\caption{Multi-Thread Concurrent Work Flow}
\label{figure:MultithreadWorkFlow}
\end{figure*}

This brings us to the second phase of this development which is to make the algorithm real-time as shown in Figure \ref{figure:MultithreadWorkFlow}. 

\subsection{Multi-Threaded Approach}
\begin{figure}
\centering
\includegraphics[width=3in]{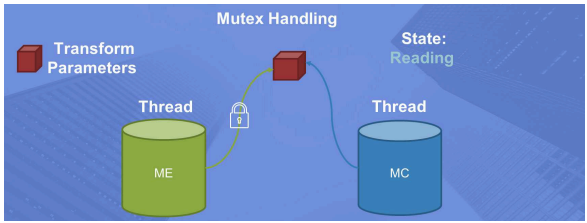}
\caption{Multi-Thread Mutex Control}
\label{figure:Mutex1}
\end{figure}

The program consists of three threads as shown in Figure \ref{figure:Mutex1}, which are capable of running independently of each other.  Parallel computing is used to separate the video stabilization algorithm  into three parts: thread 1 (motion estimation), thread 2 (motion compensation) and thread 3 (warping). Using the first thread, the motion estimation is carried out for the first twenty incoming frames (smoothing radius for motion compensation is set to ten) while the other threads wait for this process to complete. This delay is because the motion compensation and warping require smoothed corner trajectories that can be generated only after the first twenty frames are processed. From the 21$^{\rm st}$ frame, all the three threads operate simultaneously. In other words, the motion estimation between the 20$^{\rm th}$ and 21$^{\rm st}$ frame is obtained using thread 1 and the other threads compute the smoothed trajectory using the information generated from the second frame to the 21${\rm st}$ frame. The whole process continues indefinitely until the video stream ends. Since the threads themselves process the frames much faster than the frame rate, the timings of the threads eventually catch up with the timings of the frame availability.

\subsection{Motion Estimation Thread}

Once the ME thread is able to open a video file, it will keep estimating the homography matrix between each frame until it reaches the end of the video file. The motion compensation thread will be notified to run as soon as ME thread finished estimating for twenty frames.

\subsection{Motion Compensation Thread}

The motion compensation thread averages the transform parameters within the storage for the transform parameters. From now on, whenever a new frame is estimated by ME thread, the motion compensation thread will remove the oldest transform parameters, insert the latest parameters and average. It is to be reminded that the averaging at motion compensation stage has not changed from the previous implementation. The stage no longer has the whole video’s motion trajectory parameters to work with anymore. Instead, it collects and manages the data storage of the transform parameters given by motion compensation thread. 

\subsection{Image Composition Thread}

Once motion compensation produces the smoothed parameters of the latest frame, it will insert the data into the storage so that the image composition thread is able to retrieve the data and frame to display them accordingly. 

\section{Results}

\subsection{Results of Stage I}

\begin{figure*}
\centering
\includegraphics[width=3.5in]{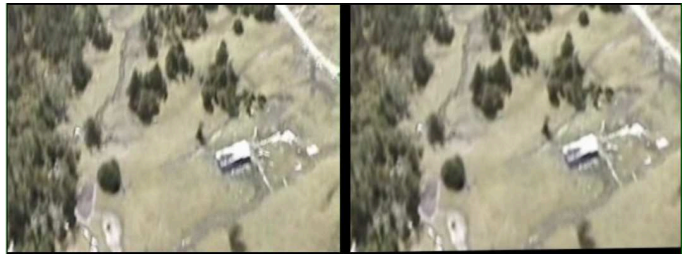}
\caption{Before and After Stabilization}
\label{figure:Before and After}
\end{figure*}

We tested the enhanced corner tracking and hybrid motion estimation system on the video obtained from in-house UAV videos and on standard benchmark videos(\cite{dong2014instantaneous}). Figure \ref{figure:Before and After} shows a screenshot of the comparison between the unstable and the stabilized video on the ‘Forest’ video from the public database.

A web demo for the ‘Forest’ video is shared below: 

\underline{http://tinyurl.com/jv6fvvu}

Compare the above result to the result obtained using rigid and similarity transformation alone:

\underline{http://tinyurl.com/gnploac} (rigid)

\underline{http://tinyurl.com/glxt65m} (similarity or partial affine)

It is evident that the hybrid motion estimation is much better in comparison to the simple rigid transformation or similarity transformation. This can be clearly witnessed after 20 seconds into the video when there are sudden camera movements.

More tests were conducted on UAV videos available as public video stabilization datasets. These tests were conducted on a desktop with a CPU clock speed of 1.7GHz. The videos used are aerial view videos taken from a UAV to simulate environment when a real UAV is used for surveillance. This algorithm is able to estimate the motion matrix of adjacent frames at a maximum speed of 100 frames per second. The same algorithm is used to estimate the motion matrix for a frame size of 640x480 at a speed of 30 frames per second. This kind of result reassures us that this algorithm is very much suited for real-time processing of motion. Below are the results of how many times rigid and affine transformations have been utilized in different videos.

\begin{table*}
\centering
\begin{tabular}{|c|c|c|} 
\hline\hline
Video & \makecell{ Number of Rigid Transformations \\ (per 20 frames)} & \makecell{Number of Similarity Transformation \\ (per 20 frames)} \\[0.5ex]
\hline
Forest & 41 & 41 \\[0.5ex]
\hline
Ground & 16 & 32 \\[0.5ex]
\hline
Bird & 21 & 9\\[0.5ex]
\hline
Girl & 12 & 10\\[0.5ex]
\hline
\end{tabular}
\caption{Number of times different motion models are used}
\label{tab:motionmodelusage}
\end{table*}

It is clear that the hybrid motion estimation uses both rigid and similarity transformations in a manner dependent on the nature of the motion. For motion behaviour that is not easy to be captured by the simple rigid mechanism, the similarity transformation offers a better model for motion behaviour. The web-demos of the other videos in the public database can be found below.

\begin{enumerate}
\item Ground: \underline{http://tinyurl.com/zfb4lo6}
\item Bird: \underline{http://tinyurl.com/jsn7g24}
\item Girl: \underline{http://tinyurl.com/hodl9ey}
\item Hippo: \underline{http://tinyurl.com/zrq8jna}
\end{enumerate}

Using video frames obtained from in-house UAV videos, we tested the hybrid motion estimation based stabilization algorithm. The captured video frames were obtained by a UAV that shows the motion of another UAV. The video frames suffer from both object motion and scene motion (due to the on-board cam motion). Below is a demo of our system. 

\underline{http://tinyurl.com/z8muwab}

From the above results, we can see that after the first few seconds, our method can better stabilize the unstable frames, especially when seen from the point of view of the UAV in the picture.  

Furthermore, we observe that although the scene changes abruptly, our system can handle them without crashing. This is a common scenario when mechanical gimbals do not provide much stability to the captured UAV videos or when there are frame rate issues or when there is a sudden mid-flight recourse taken by the UAV. All these scenarios were handled efficiently by our system, as demonstrated in the comparison video above.

\subsection{Motion Estimation Timings}
In order to ensure that our algorithm has the potential to execute in real-time, we time the duration needed to finish estimating the motion model for the publicly available videos. Most of the algorithms reviewed in the literature are either not applicable for real-time processing or needs more processing time than the proposed algorithm in this paper. Table \ref{tab:preprocesstiming} compares the processing speed of the proposed algorithm and the latest instantaneous video stabilization method in the literature. For a fair comparison, we use a notebook computer with a dual-core 1.70 GHz processor, as done in \cite{dong2014instantaneous}, using a notebook computer with a 2.5 GHz Intel dual-core CPU. Motion estimation step is the most expensive step in video stabilization. We are able to utilize optical-flow-based tracking to reduce the computational time. 

\begin{table*}
\centering
\begin{tabular}{|c|c|c|c|c|} 
\hline\hline
Video Name & Resolution & Frames & \cite{dong2014instantaneous}(fps) & Offline Method(fps)\\[0.5ex]
\hline
Forest & 320x240 & 1650 & 77.2 & 107.8\\[0.5ex]
\hline
Ground & 640x480 & 966 & 39.9 & 34.9\\[0.5ex]
\hline
Bird & 640x360 & 601 & 47.4 & 44.2\\[0.5ex]
\hline
Girl & 640x360 & 446 & 42.7 & 45.5\\[0.5ex]
\hline
\end{tabular}
\caption{Time taken to complete pre-process entire video}
\label{tab:preprocesstiming}
\end{table*}

In Table \ref{tab:preprocesstiming}, we recorded the timing for the motion estimation stage to finish computing and smoothing the global motion trajectory of the video. From the table, we can infer that the lower the resolution of the video the faster it is to calculate the motion model in between the frames. Furthermore, we speed up the algorithm by using tracking which effectively remove the need to do iterative detection of corners for each frame. This is how we achieve such an impressive timing for motion estimation.

\subsection{Results of Real-Time Implementation}

Multi-threaded implementation is possible due to the high speed of the proposed stabilization framework compared to existing works. Experiments were conducted to estimate the speed at each stage. The result was that motion compensation and image composition step combined takes much shorter time than motion estimation (ME). Hence it is concluded that ME is the bottleneck of the operation. Since through experimental results, ME can reach up to 81 frames per second on a 640x480p video, this implementation can allow the program to process frames in real-time.

 However, it has to be kept in mind that the threads execute under the same program explained in section \ref{6realtime}. This means that they are able to share resources or data with one another. This feature is perfect for the proposed algorithm because motion compensation needs to make use of the output (transform parameters) from the ME process and image composition needs to make use of the output from motion compensation as seen from Figure \ref{figure:MultithreadWorkFlow}. 

This feature can serve as a convenient way to read and write to the same resource in the memory space or it can be dangerous as data corruption can happen if the data accessibility is not handled carefully. In this program, mutex is used to facilitate resource accessibility. A demo of the in-house UAV can be viewed using the link below.

\underline{http://tinyurl.com/hel53sb}

Furthermore, with the usage of multi-threading processing, the time involved to compute and smooth the parameter trajectory is reduced significantly, as shown in Table \ref{tab:preprocesstimingrealtime}. This is because once the parameters are calculated, the motion compensation thread is able to smooth the parameter values immediately without having to wait for the computation of the parameter trajectory of the entire video to be extracted.

\begin{table*}
\centering
\begin{tabular}{|c|c|c|c|c|} 
\hline\hline
Video Name & Resolution & Frames & Offline Method(fps) & Real-Time Version\\[0.5ex]
\hline
Forest & 320x240 & 1650 & 107.8 & 126.32\\[0.5ex]
\hline
Ground & 640x480 & 966 & 34.9 & 58.93\\[0.5ex]
\hline
Bird & 640x360 & 601 & 44.2 & 72.65\\[0.5ex]
\hline
Girl & 640x360 & 446 & 45.5 & 65.36\\[0.5ex]
\hline
\end{tabular}
\caption{Comparison between offline and real-time}
\label{tab:preprocesstimingrealtime}
\end{table*}

This section contains the final video results obtained from our algorithm which are available online.
\begin{enumerate}
\item Ground: \underline{http://tinyurl.com/zfb4lo6}
\item Bird: \underline{http://tinyurl.com/jsn7g24}
\item Girl: \underline{http://tinyurl.com/hodl9ey}
\item Hippo: \underline{http://tinyurl.com/zrq8jna}
\end{enumerate}

\section{Conclusion}
This paper describes the algorithms and methods used to tackle the problem of real-time video stabilization for unmanned aerial vehicles (UAVs) videos. Due to the size and structure limitations, UAVs are highly susceptible to atmospheric turbulence, which induces jitters and makes the video unstable. It is very difficult to detect and track targets of interest in such unstable videos. Therefore, it is necessary to design a real-time video stabilization algorithm for UAVs as a pre-processing module for higher-level vision tasks, such as object detection and tracking. 
To achieve the above goal, two main components were designed as part of the proposed video stabilization algorithm: (1) By designing an appropriate motion model for the global motion of the UAVs, the proposed stabilization mechanism avoids the necessity of estimating the most general motion model, projective transformation, and considers simpler motion models like the similarity and rigid transformation; (2) In order to achieve high processing speeds, an optical flow based motion estimation is proposed to replace the conventional tracking and matching algorithms used by state-of-the-art video stabilization methods. These novel ideas resulted in a real-time stabilization algorithm. A demo of the stabilized videos can be accessed using the web links provided in this paper. 

Our method will not yield excellent results for videos that has large and gradual scene changes. This will cause the frames to be warped in such a way to resist the changes resulting in large unfilled areas in the frames. Future work will be carried out to resolve this issue.

% \section{Section title}
% \label{sec:1}
% % Citation of \citet{RefJ}.
% \subsection{Subsection title}
% \label{sec:2}
% as required. Don't forget to give each section
% and subsection a unique label (see Sect.~\ref{sec:1}).
% \paragraph{Paragraph headings} Use paragraph headings as needed.
% \begin{equation}
% a^2+b^2=c^2
% \end{equation}

% % For one-column wide figures use

% %
% % For tables use
% \begin{table}[t]
% % table caption is above the table
% \caption{Please write your table caption here}
% \centering
% \label{tab:1}       % Give a unique label
% % For LaTeX tables use
% \begin{tabular}{lll}
% \hline\noalign{\smallskip}
% first & second & third  \\
% \noalign{\smallskip}\hline\noalign{\smallskip}
% number & number & number \\
% number & number & number \\
% \noalign{\smallskip}\hline
% \end{tabular}
% \end{table}

%\begin{acknowledgements}
%If you'd like to thank anyone, place your comments here
%and remove the percent signs.
%\end{acknowledgements}

% BibTeX users please use
%\bibliographystyle{spbasic}
 %\bibliography{VIDSTAB.bbl}   % name your BibTeX data base

\begin{thebibliography}{15}
\providecommand{\natexlab}[1]{#1}
\providecommand{\url}[1]{{#1}}
\providecommand{\urlprefix}{URL }
\expandafter\ifx\csname urlstyle\endcsname\relax
  \providecommand{\doi}[1]{DOI~\discretionary{}{}{}#1}\else
  \providecommand{\doi}{DOI~\discretionary{}{}{}\begingroup
  \urlstyle{rm}\Url}\fi
\providecommand{\eprint}[2][]{\url{#2}}

\bibitem[{Adelson et~al(1984)Adelson, Anderson, Bergen, Burt, and
  Ogden}]{adelson1984pyramid}
Adelson EH, Anderson CH, Bergen JR, Burt PJ, Ogden JM (1984) Pyramid methods in
  image processing. RCA engineer 29(6):33--41

\bibitem[{Andoni and Indyk(2006)}]{andoni2006near}
Andoni A, Indyk P (2006) Near-optimal hashing algorithms for approximate
  nearest neighbor in high dimensions. In: 2006 47th Annual IEEE Symposium on
  Foundations of Computer Science (FOCS'06), IEEE, pp 459--468

\bibitem[{Battiato et~al(2007)Battiato, Gallo, Puglisi, and
  Scellato}]{battiato2007sift}
Battiato S, Gallo G, Puglisi G, Scellato S (2007) Sift features tracking for
  video stabilization. In: Image Analysis and Processing, 2007. ICIAP 2007.
  14th International Conference on, IEEE, pp 825--830

\bibitem[{Bouguet(2001)}]{bouguet2001pyramidal}
Bouguet JY (2001) Pyramidal implementation of the affine lucas kanade feature
  tracker description of the algorithm. Intel Corporation 5(1-10):4

\bibitem[{Dong et~al(2014)Dong, Xia, Yu, Su, and Hou}]{dong2014instantaneous}
Dong J, Xia Y, Yu Q, Su A, Hou W (2014) Instantaneous video stabilization for
  unmanned aerial vehicles. Journal of Electronic Imaging
  23(1):013,002--013,002

\bibitem[{Fleet and Weiss(2006)}]{fleet2006optical}
Fleet D, Weiss Y (2006) Optical flow estimation. In: Handbook of mathematical
  models in computer vision, Springer, pp 237--257

\bibitem[{Harris and Stephens(1988)}]{harris1988combined}
Harris C, Stephens M (1988) A combined corner and edge detector. In: Alvey
  vision conference, Citeseer, vol~15, p~50

\bibitem[{Ho(2014)}]{nghiaho2014simple}
Ho N (2014) Simple video stabilization using opencv.
  \urlprefix\url{http://nghiaho.com/?p=2093}

\bibitem[{Lucas et~al(1981)Lucas, Kanade et~al}]{lucas1981iterative}
Lucas BD, Kanade T, et~al (1981) An iterative image registration technique with
  an application to stereo vision. In: IJCAI, vol~81, pp 674--679

\bibitem[{Pilu(2004)}]{pilu2004video}
Pilu M (2004) Video stabilization as a variational problem and numerical
  solution with the viterbi method. In: IEEE Computer Society Conference on
  Computer Vision and Pattern Recognition

\bibitem[{Shen et~al(2009)Shen, Pan, Cheng, and Yu}]{shen2009fast}
Shen H, Pan Q, Cheng Y, Yu Y (2009) Fast video stabilization algorithm for uav.
  In: Intelligent Computing and Intelligent Systems, 2009. ICIS 2009. IEEE
  International Conference on, IEEE, vol~4, pp 542--546

\bibitem[{Shi and Tomasi(1994)}]{shi1994good}
Shi J, Tomasi C (1994) Good features to track. In: Computer Vision and Pattern
  Recognition, 1994. Proceedings CVPR'94., 1994 IEEE Computer Society
  Conference on, IEEE, pp 593--600

\bibitem[{Vazquez and Chang(2009)}]{vazquez2009real}
Vazquez M, Chang C (2009) Real-time video smoothing for small rc helicopters.
  In: Systems, Man and Cybernetics, 2009. SMC 2009. IEEE International
  Conference on, IEEE, pp 4019--4024

\bibitem[{Wang et~al(2011)Wang, Hou, Leman, and Chang}]{wang2011real}
Wang Y, Hou Z, Leman K, Chang R (2011) Real-time video stabilization for
  unmanned aerial vehicles. In: MVA, pp 336--339

\bibitem[{Yang et~al(2006)Yang, Schonfeld, Chen, and Mohamed}]{yang2006online}
Yang J, Schonfeld D, Chen C, Mohamed M (2006) Online video stabilization based
  on particle filters. In: 2006 International Conference on Image Processing,
  IEEE, pp 1545--1548

\end{thebibliography}

\end{document}